# TraffNet: Learning Causality of Traffic Generation for What-if Prediction

Ming Xu, Qiang Ai, Ruimin Li, Yunyi Ma, Geqi Qi, Xiangfu Meng, Haibo Jin

***Abstract*—Real-time what-if traffic prediction is crucial for decision making in intelligent traffic management and control. Although current deep learning methods demonstrate significant advantages in traffic prediction, they are powerless in what-if traffic prediction due to their nature of correla-tion-based. Here, we present a simple deep learning framework called TraffNet that learns the mechanisms of traffic generation for what-if pre-diction from vehicle trajectory data. First, we use a heterogeneous graph to represent the road network, allowing the model to incorporate causal features of traffic flows, such as Origin-Destination (OD) demands and routes. Next, we propose a method for learning segment representations, which models the process of assigning OD demands onto the road network. The learned segment represen-tations effectively encapsulate the intricate causes of traffic generation, facilitating downstream what-if traffic prediction. Finally, we conduct experiments on synthetic datasets to evaluate the effectiveness of TraffNet. The code and datasets of TraffNet is available at https://github.com/iCityLab/TraffNet.**

## I. INTRODUCTION

With the flourishing development of sensing, computing, and communication technologies, researchers are com-mitted to developing the next generation of intelligent transportation systems with real-time decision-making capabilities to enhance the efficiency of traffic management. During the real-time decision-making process, numerous what-if questions may arise. For instance: “what if a spe-cific road segment were subjected to traffic restrictions?”, “What if a traffic accident occurs on an arterial road?” or “What if the green light cycle of a particular traffic signal were extended by 10 seconds?” Answering these questions requires a causal short-term traffic forecasting method.

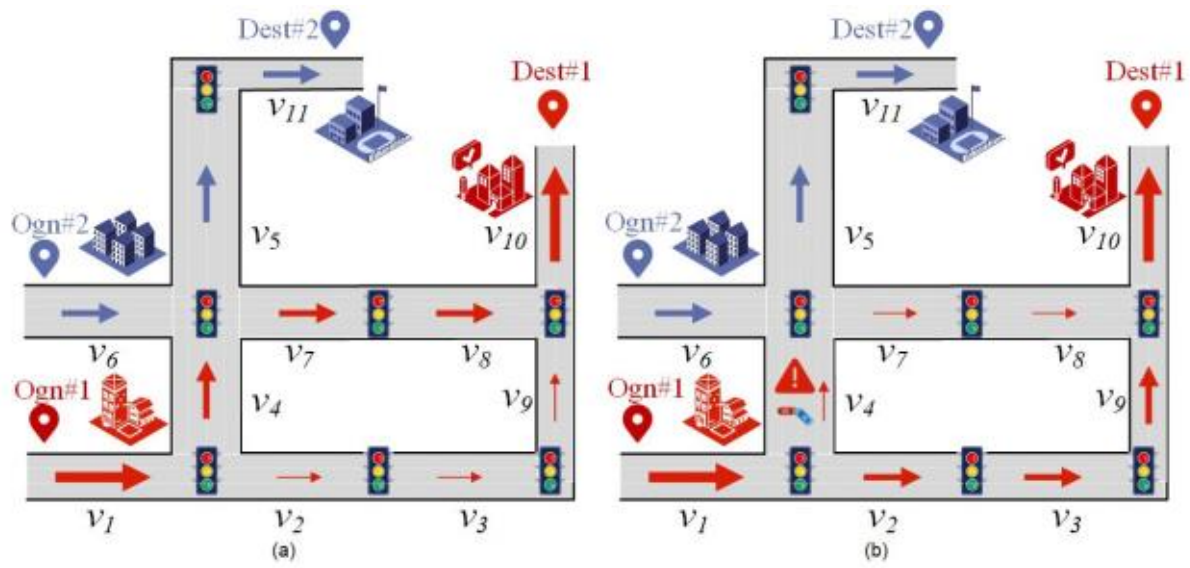

Fig. 1. An example of how OD demands and route information affects the spatial correlation. (a) Daily traffic assignment. (b) Traffic assignment during a traffic accident. A thick arrow indicates a large traffic flow.

Existing traffic forecasting methods can be classified into two categories: knowledge-based methods and data-driven methods. The knowledge-based methods utilize a traffic assignment model to evaluate medium-term or long-term traffic situations. The traffic assignment assumes that Origin-Destination (OD) demands representing volumes from origins to destinations within a specific time interval are available, and the model estimates the volumes on each segment by assigning the OD demands onto the network. However, traffic assignment contains many idealized and strict assumptions. For instance, in user equilibrium traffic assignment [37], each individual driver chooses the route with the minimal time cost for themselves, resulting in an equilibrium network state where no one can improve their travel time by changing routes. During the whole calcula-tion process of equilibrium, the model does not need to transmit any data with the real traffic system. Consequently, while knowledge-based methods model the mechanisms of traffic generation and are applicable for intervention in-ferences in mid-to-long-term traffic planning, they are unable to synchronize with the physical traffic system, making them unsuitable for real-time decision-making.

The typical representative of data-driven methods is deep learning. Recently, graph deep learning (GDL), such as graph convolution network [1], graph embedding [2], and graph attention [3], has gained significant attention and demonstrated a great advantage in traffic prediction tasks. However, the essence of deep learning models for traffic prediction lies in uncovering the correlations within traffic data, rather than intricately modeling the mechanisms of traffic generation. This means that these models cannot make reliable extrapolations on scenarios that do not produce similar

Corresponding author: Ming Xu.
Ming Xu is with the Software College, Liaoning Technical University, Huludao, Liaoning, 125100, China (e-mail: xum.2016@tsinghua.org.cn)
Qiang Ai is with the Software College, Liaoning Technical University, Huludao, Liaoning, 125100, China (e-mail: qiang.ai@outlook.com)
Ruimin Li is with the Department of Civil Engineering, Tsinghua University, Beijing, 100084 China (e-mail: lrmin@tsinghua.edu.cn).
Yunyi Ma is with the Software College, Liaoning Technical University, Huludao, Liaoning, 125100, China (e-mail: yunyima64@gmail.com)
Geqi Qi is with the Key Laboratory of Transport Industry of Big Data Application Technologies for Comprehensive Transport, Beijing Jiaotong University, Beijing, 100091, China (e-mail: gqqi@bjtu.edu.cn).
Xiangfu Meng is with the School of Electronic and Information Engineering, Liaoning Technical University, Huludao, Liaoning, 125100, China (e-mail: marxi@126.com)
Haibo Jin is with the Software College, Liaoning Technical University, Huludao, Liaoning, 125100, China (e-mail: jinhaibo@lntu.edu.cn)

historical data and therefore do not support what-if traffic prediction. To illustrate this point, let us consider the small road network in Figure 1, which contains two OD demands $\langle v_1, v_{10} \rangle$ and $\langle v_6, v_{11} \rangle$, with one and two associated paths, respectively. Figure 1-a depicts the daily traffic conditions. When a traffic incident occurs in segment $v_4$, numerous vehicles of $\langle v_1, v_{10} \rangle$ choose the secondary path instead of the primary path, as illustrated in Figure 1-b. This leads to a significant reduction in traffic on segment $v_4$, while a sudden surge on segments $v_2$, $v_3$ and $v_9$. It can be seen that the spatial dependencies between road segments is determined by OD demands and route selection. However, GDL approaches hypothesize that adjacent road segments have similar traffic changes, which does not always match reality. In this example, $v_4$ and $v_9$ can impact each other despite their distance, since $v_4$ and $v_9$ share the same OD. For the same reason, although $v_4$ and $v_5$ are neighboring, $v_4$ is independent of $v_5$. Existing GDL methods cannot make accurate predictions in this case due to the spurious spatial dependencies. While OD demand information plays a vital role in knowledge-based methods, it cannot be explicitly represented in the graph model of road network. As such, GDL approaches cannot take advantage of OD demand information to improve performance.

In summary, existing traffic forecasting methods, whether knowledge-based or data-driven, are incapable of being applied to real-time intervention inference. To solve this problem, we propose a novel deep learning framework called TraffNet, which learns the mechanisms of traffic generation from a significant amount of vehicle trajectory data. The contributions of this paper are summarized as follows.

1) We use a heterogeneous graph to represent the road network instead of a typical graph, allowing the model to incorporate causes of traffic generation, such as OD demands and route selection.

2) We design TraffNet - a novel deep learning architecture on a heterogeneous graph for short-term what-if traffic prediction. TraffNet models the process of assigning OD demands onto the road network to learn representations of road segments. The obtained segment representations encode the causal features of traffic flow, allowing for accurate matching of path-level spatial dependencies.

3) We conducted experiments on synthetic SUMO datasets to evaluate the effectiveness of TraffNet in what-if traffic prediction.

To the best of our knowledge, this is the first work to propose a deep learning architecture that learns the causality of traffic flow for what-if traffic prediction.

## II. RELATED WORK.

### A. Causal learning for Traffic-related tasks

Causal machine learning [4], which aims to discover causal relations and estimate effects from observational data to ensure the generalizability, interpretability, and fairness, has been widely used in computer vision [5], nature language process [6], recommendation systems [7], urban simulation [8], social networks [9], medical field [10], and so on. When utilizing causal models for traffic prediction, the challenges emerge from the intricate traffic causality resulting from the topological relationships within the graph. Molavipour et al. [11] present a method for inferring causal effects in traffic network by estimating the impact of traffic flow between nodes to uncover causal relations. Gao et al. [12] proposed a method for uncovering causal knowledge within traffic flow, integrating it with graph diffusion mechanisms and temporal convolutional networks to improve traffic predictions. Li et al. [13] mine the flow transitions between upstream and downstream segments from historical vehicle trajectories, which are then integrated into a graph neural network for traffic prediction. However, these studies assume that OD demands are unobservable, essentially overlooking the root causes for traffic generation and making it exceedingly challenging to identify correlations between distinct paths.

Several studies have incorporated OD demand information into traffic tasks. He et al. [14] design an OD-enhanced GCN for metro passenger flow prediction, converting OD information into a probability transition matrix used as a supplement adjacency matrix in GCN to describe dynamic station correlations. However, it lacks the feature representation of OD demands and route learning capabilities, limiting its applicability to what-if prediction. To the best of our knowledge, CRRank [15] was the first work that integrates OD demands using heterogeneous graphs in road network analysis. However, CRRank relies on heuristic rules and is not a parametric learnable model, and its accuracy still needs improvement.

### B. GNN for Traffic-related tasks

Traffic-related tasks are a critical topic in the field of spatio-temporal data mining and intelligent transportation. In recent years, graph deep learning methods have achieved state-of-the-art performance in fine-grained (e.g., road segment level) traffic prediction, thanks to their accurate modeling of spatio-temporal correlations in traffic data. Previous studies [16-19] use two separate modules to capture temporal and spatial dependencies. Song et al. [20] attempt to incorporate spatial and temporal blocks together to capture simultaneously localized spatial-temporal correlations. Zhang et al. [21] design a hierarchically structured graph neural architecture to capture cross-region inter-dependencies and a multi-scale attention network to model multi-resolution temporal dynamics. Li et al. [22] introduce a novel spatial-temporal fusion graph module to capture long-range spatio-temporal dependencies synchronously. Cui et al. [23] combines traffic graph convolutional and LSTM to learn spatio-temporal dependencies. Although these methods have achieved a certain degree of improvement, their nature based on correlation prevents them from making what-if prediction.

### C. Heterogeneous Graph Learning

Heterogeneous graphs have been widely used in, text mining [26], identification [29], recommendation systems [24,25] and other fields [30, 31]. To support these applications, heterogeneous graph representation learning (HGRL) is proposed. HGRL aims to learn a function that embeds nodes in the heterogeneous graph into a low-dimensional Euclidean space while preserving both structural and semantic information. However, since a heterogeneous graph is closely related to a specific domain, existing HGRL methods cannot directly solve the problems faced in this paper.

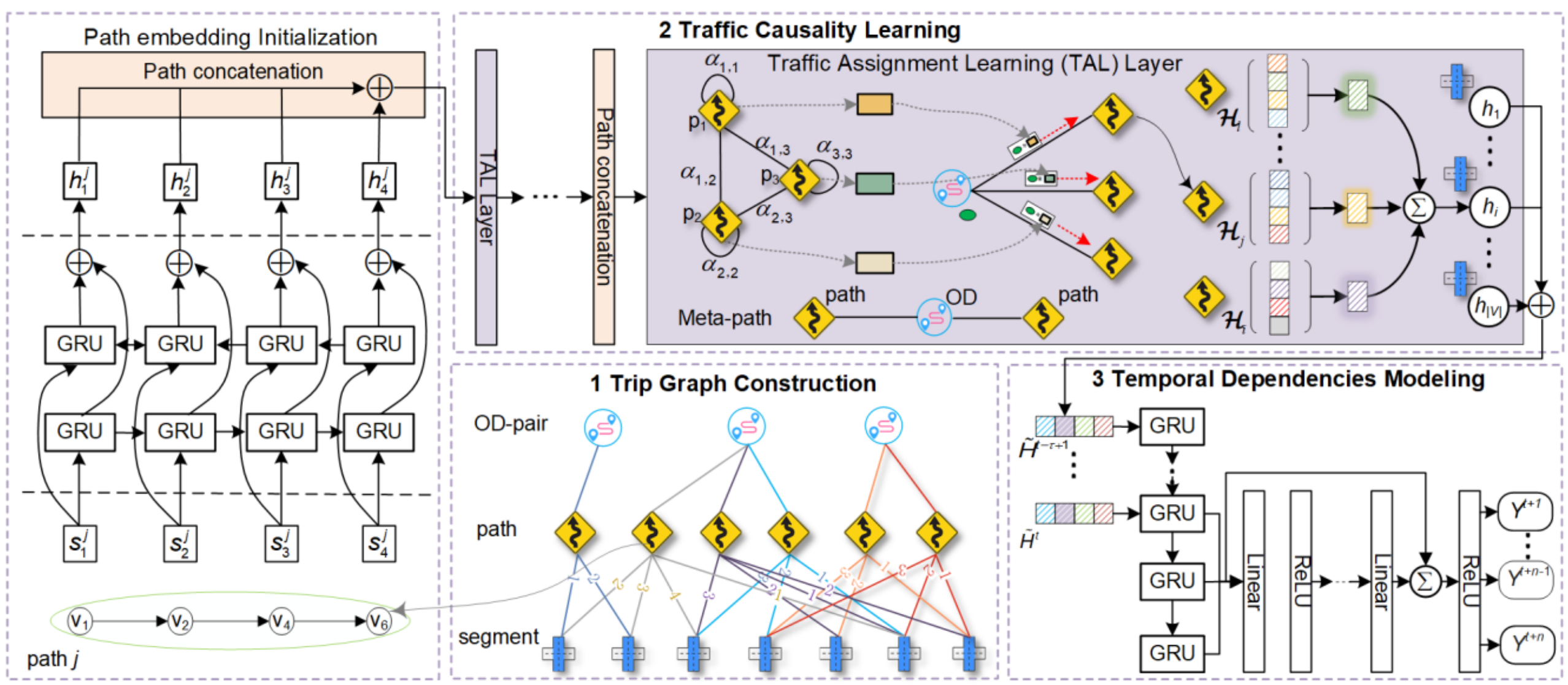

**Fig. 2.** Architecture of TraffNet.

## III. Methodology

### A. Problem Formalization

The definition used in this paper are given as follow.

**Road network**: A road network is a directed graph $G$ ($V$, $E$), where $V$ is the set of nodes; $E$ is the set of directed edges; An edge $\langle v_i, v_j \rangle$ indicates that node $v_i$ is the incoming neighbor of node $v_j$. Each node $v_i$ has an attribute vector $a_i$. $A$ denotes the set of attributes of all nodes in network $G$. In this paper, a node specifically refers to a segment.

**OD demands**: Given a road network $G(V, E)$, OD demands $M^t$ during a specific time interval $t$ can be represented as a matrix $\{m_{i,j}\}$, where each element $m_{i,j}$ is the number of vehicles departing from $v_i$ to $v_j$.

**Path**: A path $p$ is a node sequence $\langle v_1, v_2, \cdots, v_{|p|} \rangle$ including all the nodes traversed by a trip from an OD demand. Here, $/p/$ is the number of nodes in $p$. Obviously, an OD demand can be associated with one or multiple paths.

**Traffic volume**: Given a road network, traffic volumes $X^t$ can be represented as a traffic vector $\{x_i^t\}$, where each element $x_i^t$ represents the number of vehicles that have passed through node $v_i$ during the time interval $t$.

**Traffic prediction**: For a road network, given OD matrices $M^{t-\tau+1:t}$, traffic volumes $X^{t-\tau+1:t}$ of $\tau$ historical time intervals before interval $t$ and attributes $A$, traffic prediction computes the traffic vectors for the next $l$ intervals using the function $f$

$$\hat{X}^{t+1:t+l} = \hat{f}_\theta(X^{t-\tau+1:t}, M^{t-\tau+1:t}, A) \tag{1}$$

where $\hat{f}_\theta$ aims to minimize the error $J$ between the evaluated and real traffic volumes.

$$\min_\theta J\left(X^{t+1:t+l}, \hat{X}^{t+1:t+L}\right) \tag{2}$$

**What-if traffic prediction.** For the function $\hat{f}_\theta$, what-if traffic prediction explores different predicted outcomes by setting different input values ($A$ or $M$).

We propose TraffNet - a novel deep learning framework on a heterogeneous graph for the what-if traffic prediction task. As shown in Fig. 2, TraffNet consists of three modules: Trip Graph Construction, Traffic Causality Learning, Temporal Dependencies Modeling. In the trip graph construction module, TraffNet extracts OD demands and routes from vehicle trajectories and integrates the road network structural properties and traffic conditions to construct a heterogeneous graph. In the traffic causality learning module, TraffNet learn a segment representation that encodes causes of traffic flows for each segment. This module comprises three components: path embedding, route selection, and road segment embedding. The temporal dependencies modeling module is used to match the temporal underlying process of traffic generation.

### B. Trip Graph Construction

We construct the trip graph by extracting OD demands and routes from vehicle trajectories. A trip Graph $TG$ can be formulated as a tripartite graph $\{M, P, V, R, R'\}$, where $M$, $P$ and $V$ are sets of three types of nodes: OD-pairs, paths, and road segments. $R$ is the set of edges linking $M$ and $P$, and an edge $\langle m, p \rangle \in R$ represents that OD-pair $m$ is associated with path $p$; $R'$ is the set of edges linking $P$ and $V$ and an edge $\langle p, v; k \rangle \in R'$ represents that segment $v$ is the k-th segment in path $p$. $m$, $p$ and $v$ have an attribute vector, respectively.

It's worth noting that the topology of the trip graph is stable if there is enough trajectory data.

### C. Traffic Causality Learning

In the traffic causality learning module, representation learning for road segments aims to acquire concise, low-dimensional representations that effectively encapsulate the complexity of causes of traffic generation. Subsequently, the learned segment representations facilitate addressing downstream traffic prediction task. This module begins with the path embedding initialization layer, and then alternately stacks Traffic Assignment Learning (TAL) layer and path concatenation layer to compute the embedding for all the segments.

**Path Embedding Initialization**. To capture the effect of traffic propagation along a path, path embeddings must encapsulate the attributes and traffic information of all segments along the path, as well as the mutual influence among these segments. Within a path, traffic flow propagates forward from upstream segments to downstream segments, while congestion propagates backward from downstream to upstream segments. The speed and direction of propagation is determined by factors such as segment capacity, traffic conditions, and congestion levels. As such, each segment is influenced by both its upstream and downstream segments. The intra-path propagation effect can be viewed as analogous to the semantic context of a word in a sentence. In natural language processing tasks, a sequence model like GRU is often used to extract context vectors. GRU employs a gate mechanism to determine which previous information should be ignored and what new information should be incorporated. To learn the path embedding, TraffNet first uses a bidirectional GRU (Bi-GRU) to learn the representation of each segment in the path, which captures propagation effect in both directions. Note that the same segment can have different representations depending on the path it belongs to. The process of segment embedding can be formalized as follow:

$$h_i^{j(f)} = GRU^{(f)}\left(o_i^j, h_{i-1}^{j(f)} | \theta^{(f)}\right) \quad (3)$$

$$h_i^{j(b)} = GRU^{(b)}\left(o_i^j, h_{i+1}^{j(b)} | \theta^{(b)}\right) \quad (4)$$

$$h_i^j = h_i^{j(f)} \oplus h_i^{j(b)} \quad (5)$$

where $o_i^j$ is the initial representation of the $i$-th segment of path $p_j$, which is obtained by concatenating its one-hot vector and features; $\theta^{(f)}$ and $\theta^{(b)}$ are the learnable parameters, $h_i^{j(f)}$ and $h_i^{j(b)}$ are the outputs of hidden states that preserve the useful information of upstream and downstream segments for the $i$-th segment. The operator "$\oplus$" represents the concatenating of two vectors. $h_i^j$ is the intra-path embedding of the $i$-th segment in path $p_j$, which encodes both forward and backward propagation effects. Then, a path concatenation layer that concatenates the intra-path representations of all of segment in each path together is used to initialize the path-node embedding. The embedding $\mathcal{H}_j$ of the path $p_j$ is obtained as:

$$\mathcal{H}_j = h_1^j \oplus h_2^j \cdots \oplus h_{|N_V(j)|}^j \quad (6)$$

where $|N_V(j)|$ is the number of segments in path $p_j$.

**TAL Layer.** TAL layer calculates segment embeddings based on current path embeddings, enabling them to encode the corresponding OD demand and the path-level spatial dependencies. TAL and PC can be stacked alternately for $n$ ($n \geqslant 1$) layers to learn traffic causality. TAL layer consists of two components: the Route Learning and Segment Message Passing. The route learning component is responsible for determining the proportion of OD demands assigned to each candidate route under dynamic traffic conditions. Drivers commonly choose routes based on factors such as road capacity, surroundings, distance, and current traffic conditions. Since the path-node embedding captures the intra-path propagation effects, these factors are already encoded in the path-node embedding, and thus the path-node embedding can be used to learn route preferences. Specifically, TraffNet uses meta-path [24] based graph attention to learn the importance of a path-node and its neighbors of the same OD, and then fuses this information to calculate the route preference. Based on the route preferences, TraffNet updates each path-node embedding to incorporate the OD demands assigned to the path. First, under the guidance of the meta-path Path-OD-Path, all the neighboring path-nodes (associated with the same OD-node) of a given path-node are obtained. Then, graph attention is employed to form the new path-node embedding, which aggregates the representation of meta-path neighbors. Specifically, attention scores $\alpha_{j,d}$ are computed to indicate the importance of $\mathcal{H}_d$ to $\mathcal{H}_j$. The embedding $\mathcal{H}_j{}'$ of path-node $p_j$ is then updated based on the aggregated information from the meta-path neighbors. The process can be formalized as:

$$\mathcal{H}_j' = \sum_{d \in N(j)} \alpha_{j,d} W' \mathcal{H}_d \quad (7)$$

$$\alpha_{j,d} = \frac{exp\left(LeakyRelu\left(a\left[W\, \mathcal{H}_j \oplus W\, \mathcal{H}_d\right]\right)\right)}{\sum_{l \in N(j)} exp\left(LeakyRelu\left(a\left[W\mathcal{H}_j \oplus W\mathcal{H}_l\right]\right)\right)} \quad (8)$$

where $a$ is a learnable parameter; $W$ and $W'$ are the shared weight matrices for linear transformation; $N(j)$ is the collection of meta-path neighbors of path-node $p_j$. Note that meta-path neighbors include itself. Then, we adopt a multi-layer fully connected network and a softmax function to compute the route preference $co_j^k$ according to the path-node embedding $\mathcal{H}_j{}'$. $co_j^k$ refers to the proportion of choosing path $p_j$ for OD-pair $m_k$. The formula is given as:

$$co_j^k = \frac{exp\left(MLP(\mathcal{H}_j{}')\right)}{\sum_{l \in N(j)} exp\left(MLP(\mathcal{H}_l{}')\right)} \quad (9)$$

Finally, we obtain an updated path-node embedding that incorporates the traffic assignment:

$$\widetilde{\mathcal{H}}_j = m_k \cdot co_j^k \cdot \mathcal{H}_j \quad (10)$$

where $m_k \cdot co_j^k$ refers to the traffic from OD demand $m_k$ assigned to path $p_j$.

Segment Message Passing component is used to obtain the updated segment embedding according to the path-node embedding. Intuitively, a segment may belong to multiple paths, and its traffic volumes are influenced by the flow propagation and merging on all associated paths. Therefore, the embedding of a segment needs to capture not only the propagation effect within a path, but also the merging effect among paths. To obtain a segment embedding, TraffNet aggregates all path embeddings related to the segment, while considering the position of the segment in each path. We propose a position-aware message passing mechanism on the path-segment bipartite subgraph, where the position of the segment in the associated paths serves as the order attribute of the path-segment edges. First, based on the edge order, we extract the intra-path information of a segment from all the associated path-node embeddings and pass it to the segment-node as messages. Then, the segment-node aggregates the received messages to obtain a segment embedding that captures the intra-path propagation effect, route selection pattern, and inter-path merging effect. The process of position-aware message passing is formulated as:

$$h_i = \sum_{j \in N(i)} \left[\widetilde{\mathcal{H}}_j\right]_k^n \quad (11)$$

where the operator $[\cdot]_k^n$ returns the $k$-th sub-vector of length $n$ of a high dimensional vector. Here, $n$ is the dimension of the intra-path embedding of the segment. $k$ is the order of path-segment edge. It is worth noting that the precision of segment representations and path representations mutually reinforce each other. Therefore, we alternately stack multiple path concatenation layers and TAL Layers to learn more accurate segment representations.

### D. Temporal Dependencies Modeling

TraffNet aims to predict future traffic volumes for each road segment based on observed OD demands and traffic conditions. The current segment embedding encodes the factors that contribute to traffic generation, such as OD demands and path-level spatial dependencies, in a low-dimensional vector. In the temporal dependencies modeling module, we use a GRU to model the time-varying process. We feed all the segment embeddings $\widetilde{H}_t$ from the previous $t$ intervals into the temporal GRU module as input and use it to obtain traffic forecasts for the next L intervals. The temporal GRU can be expressed as:

$$Z^{t+1} = GRU\left(\widetilde{H}_t, Z^t | \theta'\right) \quad (12)$$

where $Z^t$ is the hidden state in interval $t$ of the temporal GRU; $\theta'$ is the learnable parameter. $\widetilde{H}_t$ is the matrix of all the segment embeddings at $t$ intervals. The output module of TraffNet consists of a multi-layer fully connected network with a residual block:

$$\hat{Y}^{t+1:t+L} = Relu\left(Z^{t+1:t+L} W_{res} + MLP\left(Z^{t+1:t+L}\right)\right) \quad (13)$$

where $W_{res}$ is a learnable parameter matrix; $\hat{Y}^{t+1:t+L}$ is the predicted traffic volume of TraffNet from $t$+1 to $t$+$L$ intervals.

### E. Model Training

The training process can be separated into two stages: online training and offline training. In offline training, a pre-training task for route prediction can be performed on the traffic causality learning module using historical routing data. This task sets the traffic volumes of paths as pre-training labels and can be viewed as a node regression problem for a graph where label value of each node is predicted given the graph topology and the attributes of the nodes. When training TraffNet, the traffic causality learning module is initialized with pre-trained parameters, while the rest of the parameters are initialized randomly. During the traffic prediction, all parameters are fine-tuned, with both the pre-training loss and traffic prediction loss computed as mean squared error (MAE). We use the Adam optimizer for model learning.

During online training, TraffNet makes predictions at each time interval and the model is updated every $\phi$ intervals, where $\phi$ is a configurable parameter. Initially, the off-training parameters are used to initialize TraffNet. At current time interval $t$, new training samples are generated by combining information from the newly arrived real-time data and historical data. The prediction loss of TraffNet is calculated for the new sample at interval $t$. After the losses of $\phi$ intervals are accumulated, the accumulated losses are used as feedback to calculate gradients and update all parameters.

## IV. EXPERIMENT

TraffNet is implemented with PyTorch and DGL. All the experiments are conducted using a computer with one NVIDIA RTX 3090 GPU cards.

### A. Dataset

We build two synthetic datasets using traffic simulator SUMO [25] to evaluate the effectiveness of TraffNet.

Sumo-SY. The simulated road network is a region in Shenyang, consisting of 254 road segments and 86 intersections. We collected traffic data from the major roads for a period of 28 days and inferred OD demands with transCAD. These OD demands were then used to configure the simulated road network. Finally, SUMO was run to generate the dataset, which includes 29 OD-pairs and 76 paths. The time interval is set to 2 minutes.

Sumo-WI is a variant of Sumo-SY, enriched with what-if scenarios. It contains 14 days of data with 29 OD pairs and 121 paths. Sumo-WI simulates some events including accidents or traffic restrictions. During the event, we randomly select a few road segments and reduce their traffic capacity by a random percentage, causing changes in traffic flow across multiple segments over a certain period. However, the proportion of event-affected segments in the network is very low during any time interval, resulting in an imbalanced learning problem. The labels of minority event-affected segments are overwhelmed by those of the unaffected segments, which can degrade the prediction performance of the model when using the mean square error of all segments as the loss function. To address this problem, we increase the weights of affected segments on the training loss for all of baselines and TraffNet. The modified training loss function on Sumo-WI is given as follows:

$$\text{Loss} = \frac{1}{|V| \cdot L} \sum_{l=1}^{L} \left( \sum_{v_i \in \Phi} \beta \left(y_i^l - \hat{y}_i^l\right)^2 + \sum_{v_j \in \overline{\Phi}} \left(y_j^l - \hat{y}_j^l\right)^2 \right) \quad (14)$$

where $\Phi$ and $\overline{\Phi}$ are the sets of event-affected and unaffected segments, respectively; $\beta$ is the weight of event-affected segments and is set to 10. We split all the three datasets into training, validation, and test sets with ratios of 60%, 10%, and 30%.

### B. Baselines and Evaluation Metrics

The baselines include the traditional time series approaches, deep learning approaches and graph-based Spatial-Temporal models. We also compare the variants of TraffNet (TraffNet-noOd and TraffNet-noTF) to evaluate the impact of OD demands and traffic volume of segment. Specifically, (i) HA [32] uses the average of historical traffic volume for prediction; (ii) SVR [33] uses support vector machine for regression; (iii)

Table I
Performance comparison of TraffNet and other baselines on Sumo-SY

| Model | Horizon 1 | | Horizon 2 | | Horizon 3 | | Horizon 4 | | Horizon 5 | | Horizon 6 | |
|---|---|---|---|---|---|---|---|---|---|---|---|---|
| | RMSE | MAE | RMSE | MAE | RMSE | MAE | RMSE | MAE | RMSE | MAE | RMSE | MAE |
| HA | 1.5857 | 0.8350 | 1.5955 | 0.8408 | 1.6102 | 0.8493 | 1.6241 | 0.8570 | 1.6383 | 0.8648 | 1.6503 | 0.8718 |
| SVR | 1.5840 | 0.8353 | 1.5920 | 0.8402 | 1.6023 | 0.8467 | 1.6126 | 0.8526 | 1.6240 | 0.8585 | 1.6331 | 0.8639 |
| GRU | 1.5770 | 0.8325 | 1.5901 | 0.8420 | 1.5993 | 0.8455 | 1.6075 | 0.8495 | 1.6163 | 0.8547 | 1.6243 | 0.8614 |
| T-GCN | 1.5833 | 0.8388 | 1.5931 | 0.8466 | 1.6048 | 0.8562 | 1.6154 | 0.8624 | 1.6241 | 0.8672 | 1.6353 | 0.8735 |
| STGCN | 1.5745 | 0.8421 | 1.5813 | 0.8471 | 1.5891 | 0.8528 | 1.5974 | 0.8576 | 1.6063 | 0.8638 | 1.6143 | 0.8702 |
| DCRNN | 1.5785 | 0.8394 | 1.5842 | 0.8440 | 1.5989 | 0.8499 | 1.6045 | 0.8550 | 1.6123 | 0.8600 | 1.6270 | 0.8672 |
| GraphWaveNet | 1.5658 | 0.8380 | 1.5743 | 0.8438 | 1.5836 | 0.8491 | 1.5912 | 0.8548 | 1.6004 | 0.8607 | 1.6076 | 0.8667 |
| TraffNet-noOD | 1.3944 | 0.7371 | 1.4460 | 0.7608 | 1.4958 | 0.7864 | 1.5287 | 0.8041 | 1.5487 | 0.8162 | 1.5602 | 0.8249 |
| TraffNet-noTF | 1.3352 | 0.6907 | 1.3795 | 0.7236 | 1.4434 | 0.7610 | **1.4974** | 0.7916 | **1.5306** | 0.8103 | **1.5521** | 0.8230 |
| TraffNet | **1.3170** | **0.6825** | **1.3692** | **0.7174** | **1.4408** | **0.7564** | 1.4997 | **0.7882** | 1.5363 | **0.8079** | 1.5589 | **0.8202** |

Table II
Performance comparison of TraffNet and other baselines on Sumo-WI

| Model | Horizon 1 | | Horizon 2 | | Horizon 3 | | Horizon 4 | | Horizon 5 | | Horizon 6 | |
|---|---|---|---|---|---|---|---|---|---|---|---|---|
| | RMSE | MAE | RMSE | MAE | RMSE | MAE | RMSE | MAE | RMSE | MAE | RMSE | MAE |
| HA | 1.6592 | 0.8856 | 1.6714 | 0.8921 | 1.6884 | 0.9012 | 1.7060 | 0.9105 | 1.7228 | 0.9194 | 1.7374 | 0.9271 |
| SVR | 1.6594 | 0.8939 | 1.6684 | 0.9004 | 1.6827 | 0.9099 | 1.6973 | 0.9184 | 1.7114 | 0.9270 | 1.7238 | 0.9340 |
| GRU | 1.7486 | 0.9255 | 1.7545 | 0.9369 | 1.7711 | 0.9526 | 1.7815 | 0.9430 | 1.7759 | 0.9552 | 1.7966 | 0.9715 |
| T-GCN | 1.7283 | 0.9360 | 1.7384 | 0.9263 | 1.7377 | 0.9306 | 1.7626 | 0.9422 | 1.7600 | 0.9554 | 1.7786 | 0.9690 |
| STGCN | 1.7038 | 0.9363 | 1.7134 | 0.9425 | 1.7189 | 0.9448 | 1.7279 | 0.9388 | 1.7347 | 0.9477 | 1.7399 | 0.9518 |
| DCRNN | 1.7198 | 0.9366 | 1.7225 | 0.9409 | 1.7256 | 0.9452 | 1.7388 | 0.9470 | 1.7501 | 0.9520 | 1.7599 | 0.9589 |
| GraphWaveNet | 1.7160 | 0.9349 | 1.7214 | 0.9432 | 1.7292 | 0.9454 | 1.7425 | 0.9532 | 1.7447 | 0.9515 | 1.7500 | 0.9595 |
| TraffNet-noOD | 1.5266 | 0.8047 | 1.6016 | 0.8485 | 1.6793 | 0.8815 | 1.6958 | 0.9081 | 1.7072 | 0.9253 | 1.7299 | 0.9064 |
| TraffNet-noTF | 1.4673 | 0.7493 | 1.5329 | 0.8011 | 1.5970 | 0.8402 | 1.6507 | 0.8737 | 1.6790 | 0.8871 | 1.6934 | **0.8930** |
| TraffNet | **1.4153** | **0.7267** | **1.5122** | **0.7892** | **1.5898** | **0.8338** | **1.6419** | **0.8671** | **1.6727** | **0.8840** | **1.6790** | 0.8953 |

GRU [34] is a variant of RNN widely used in time series prediction; (iv) STGCN [18] is a graph neural network which models spatio-temporal correlation via complete convolutional structures; (v) T-GCN [16] captures the spatial-temporal correlations with GCN and GRU; (vi) DCRNN [17] integrates the diffusion convolution operator with graph neural networks to predict traffic flow; (vii) GraphWaveNet [35] captures spatial-temporal dependencies by combining dilated casual convolution with graph convolution; (viii) TraffNet-noOD is a variant of TraffNet, which sets all of the OD demands as 1 to ignore the impact of OD demands; (ix) TraffNet-noTF is a variant of TraffNet, which sets all of dynamic traffic features (volume and average speed) as 0 to ignore their impact.

We adopt mean absolute error (MAE) and rooted mean square error (RMSE) to evaluate our experimental results.

### C. Setting

The hyper-parameters for TraffNet are as follows: The Path Embedding adopts a 2-layer bidirectional GRU architecture. The hidden layer dimension is 400 on Sumo-WI, 280 on Sumo-SY. The route learning module consists of a 2-attention-head GAT with 1024-dimensional hidden layers and one 7-layer FCN. The temporal dependency modeling module consists of a 2-layer GRU with 1200 hidden dimensions, and a multi-layer FCN with residual blocks. The number of FCN layers is 12 for Sumo-WI and 10 for Sumo-SY.

### D. Experimental Results

Table I and Table II show the performance of TraffNet and the baselines in Sumo-SY and Sumo-WI, respectively. we have the following observations: (1) TraffNet achieves the best performance in almost all prediction horizons, with particularly noticeable improvements in short-term prediction. In the first prediction horizon, the improvement is 15.88% on RMSE and 18.01% on MAE in Sumo-SY, while the improvement is 14.70% on RMSE and 17.94% on MAE in Sumo-WI. As the prediction horizon increases, the accuracy of TraffNet decreases more rapidly than other baselines. In the 6th horizon, TraffNet performs slightly better than baselines in Sumo-SY and Sumo-WI. This suggests that earlier OD demands are less useful in traffic evaluation, possibly because associated trips have already ended. (2) HA and SVR achieve similar performance to the deep learning baselines in Sumo-SY, while outperforming them in Sumo-WI. This may potentially be linked to the noise in the data. Due to the short time intervals of simulated data, traffic time series data often exhibit large fluctuations and contain a significant amount of random noise. Over-parameterized deep learning models have strong representation capabilities but are more susceptible to noise compared to HA and SVR models. This makes deep learning models more prone to over-fitting and reduces their generalization performance. In terms of Sumo-WI, simulated accidents or traffic restrictions can result in unexpected traffic changes that weaken the correlation between neighboring segments. Therefore, graph deep learning models that rely on the correlation to make predictions exhibit poor performance in Sumo-WI. (3) TraffNet-noOD has a higher error than TraffNet, indicating that OD information makes a significant contribution to traffic evaluation. The similar performance of TraffNet-noTF to TraffNet suggests that traffic feature may not contribute to traffic prediction as significantly as OD information. Fig. 3. compare the performance of TraffNet and other models on a specific road segment in the first prediction horizon in Sumo-SY and Sumo-WI, respectively. The results indicate that TraffNet has the best performance.

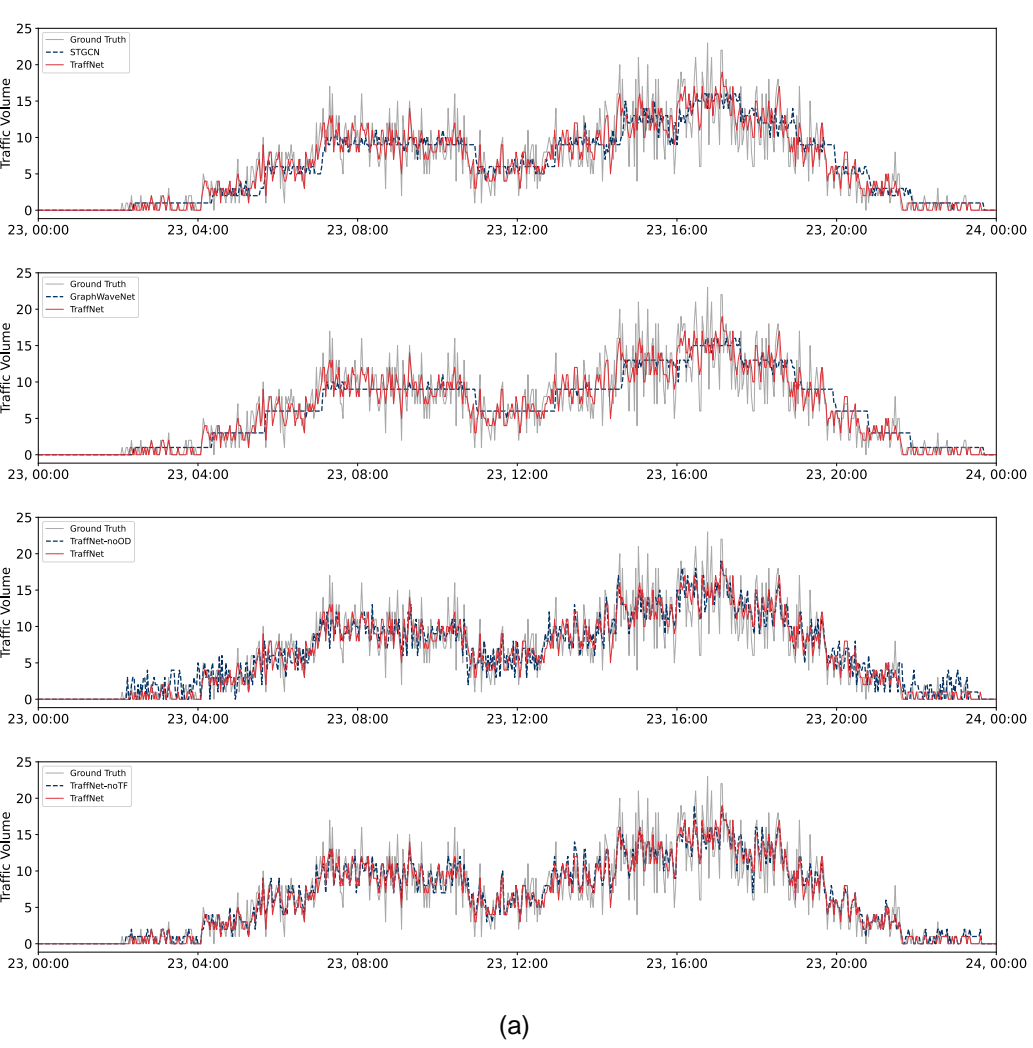

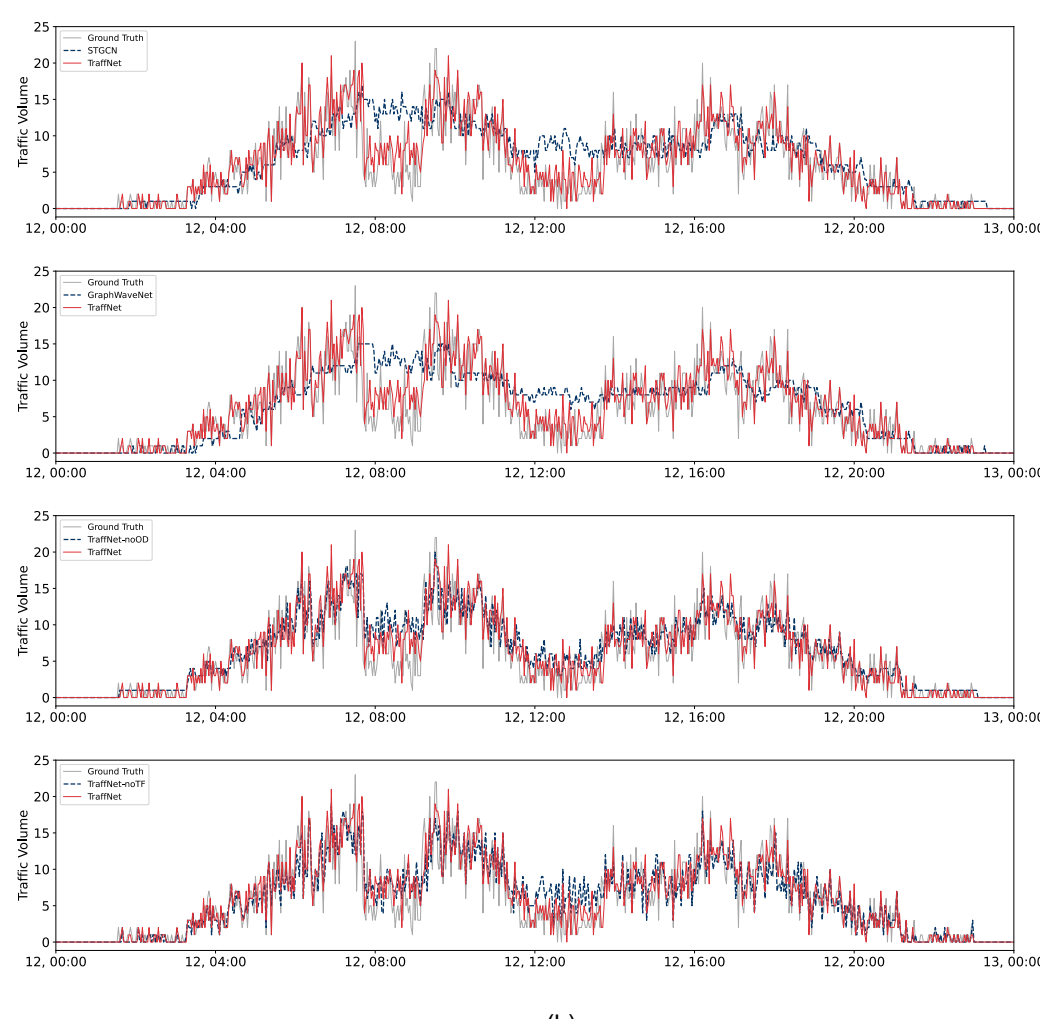

**Fig. 3.** Performance comparison of TraffNet and the top two performing baselines and its variants on a specific road segment in (a) Sumo-SY and (b) Sumo-WI.

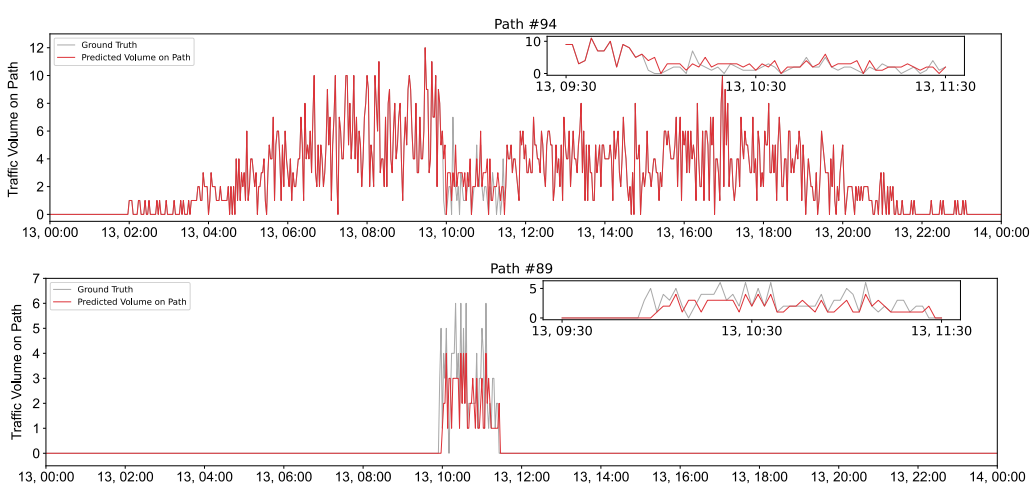

**Fig. 4.** Prediction results of the route learning module in a traffic restriction simulation case.

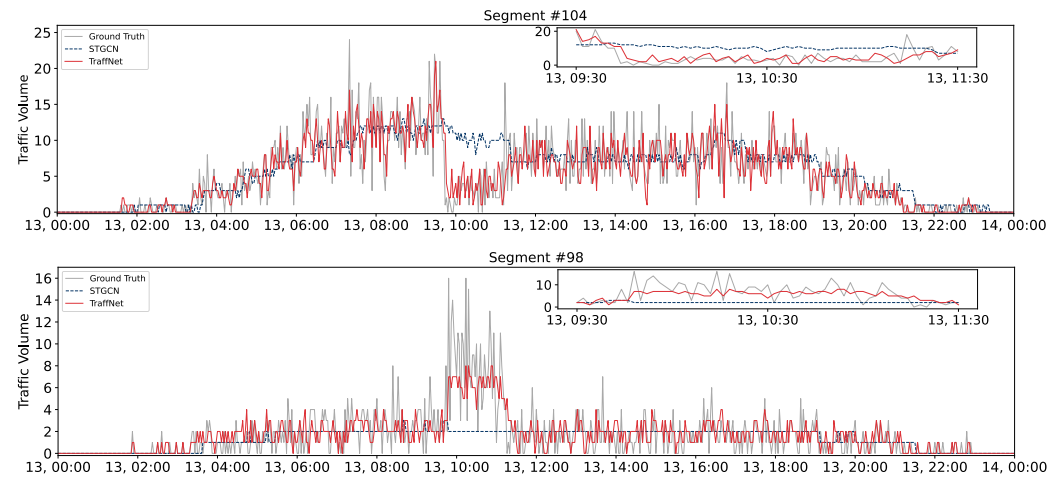

**Fig. 5.** Comparison of TraffNet and the best performing baseline on event-affected segments in Sumo-WI.

In this section, we demonstrate the effectiveness of the route learning module and the capability of TraffNet to adapt to unexpected traffic variation. After pre-trained in Sumo-WI, the route learning module achieves an overall accuracy of 76.49%. We study a traffic accident case in Sumo-WI. This accident causes a significant reduction in the traffic speed of Segment #104 to 10% between 10:00 am and 11:30 am on day 13, leading to most drivers changing their routes from Path #94 to Path #89. Fig. 4. shows that the route learning module accurately predicts the route changes. As a segment of Path #89, the traffic volume on Segment #98 increases during this accident. Fig.5. displays the predicted results of TraffNet and STGCN for Segments #104 and #98 during this period. We can see that TraffNet provides a more accurate prediction than STGCN during unexpected traffic variation periods. This verifies that TraffNet has learned the causal relation between OD demands and traffic volumes, enabling it to predict the association changes between paths.

## V. Conclusion and Feature Work

This paper proposes a deep learning model called TraffNet, which learns causality of traffic flow from vehicle trajectories for what-if traffic prediction. First, a trip graph is introduced to uniformly represent OD demands, routes, road segments, and their relationships. Based on the trip graph, we proposed a segment embedding method that model the process of traffic assignment to learn the segment representation. The learned segment embedding encode the causes of traffic generation, which are then integrated with a temporal dependency learning module to make traffic prediction. The effectiveness of TraffNet was evaluated through experiments on synthetic datasets. The results show that TraffNet achieves the best performance compared with all the baselines, with a particularly significant advantage in predicting traffic volume on event-affected segment. The reason why TraffNet is competent for what-if analysis is that we integrate domain knowledge of traffic assignment into the design of the deep learning model, which enables the model to accurately capture causal relation between OD demands and traffic volumes.

Real-time decision making requires a what-if simulation to select the best behavior and avoid any undesirable consequences from direct execution in the physical system. Since there are potentially many what-if scenarios in the road network that need to be simulated to generate various types of decisions, current versions of TraffNet can only support a few simple what-if scenarios, such as short-term traffic restriction simulations for a certain road segment. However, we believe that an extension to TraffNet will support more what-if simulations, such as traffic signal phase control simulations and large truck restriction simulations. To achieve this, the model should consider more refined features, such as the OD demand of different vehicle types and intersection signals. Additionally, it should also make accurate predictions of OD demands to support the scenarios where real-time OD demands are unavailable, which is left as our future work.

# References

[1] Zhang S, Tong H, Xu J, et al. Graph convolutional networks: a comprehensive review[J]. Computational Social Networks, 2019, 6(1): 1-23.

[2] Cai H, Zheng V W, Chang K C C. A comprehensive survey of graph embedding: Problems, techniques, and applications[J]. IEEE transactions on knowledge and data engineering, 2018, 30(9): 1616-1637.

[3] Veličković P, Cucurull G, Casanova A, et al. Graph attention networks. ICLR, 2018, pp. 1-12.

[4] Kaddour J, Lynch A, Liu Q, et al. Causal machine learning: A survey and open problems[J]. arXiv preprint arXiv:2206.15475, 2022.

[5] Hartford J, Lewis G, Leyton-Brown K, et al. Deep IV: A flexible approach for counterfactual prediction[C]. PMLR, 2017: 1414-1423.

[6] Vig J, Gehrmann S, Belinkov Y, et al. Investigating gender bias in language models using causal mediation analysis[J]. Advances in neural information processing systems, 2020, 33: 12388-12401.

[7] Wang W, Zhang Y, Li H, et al. Causal recommendation: Progresses and future directions[C] SIGIR. 2023: 3432-3435.

[8] McDuff D, Song Y, Lee J, et al. Causalcity: Complex simulations with agency for causal discovery and reasoning[C]. PMLR, 2022: 559-575.

[9] Guo R, Li J, Liu H. Learning individual causal effects from networked observational data[C]. WSDM. 2020: 232-240.

[10] Kaddour J, Zhu Y, Liu Q, et al. Causal effect inference for structured treatments[J]. Advances in Neural Information Processing Systems, 2021, 34: 24841-24854.

[11] Molavipour S, Bassi G, Čičić M, et al. Causality graph of vehicular traffic flow[J]. arXiv preprint arXiv:2011.11323, 2020.

[12] Gao T, Marques R K, Yu L. GT-CausIn: a novel causal-based insight for traffic prediction[J]. arXiv preprint arXiv:2212.05782, 2022.

[13] Li M, Tong P, Li M, et al. Traffic flow prediction with vehicle trajectories. AAAI. 2021, 35(1): 294-302.

[14] He C, Wang H, Jiang X, et al. Dyna-PTM: OD-enhanced GCN for Metro Passenger Flow Prediction. IJCNN. IEEE, 2021: 1-9.

[15] Xu M, Wu J, Liu M, et al. Discovery of critical nodes in road networks through mining from vehicle trajectories. IEEE Transactions on Intelligent Transportation Systems, 2018, 20(2): 583-593.

[16] Zhao L, Song Y, Zhang C, et al. T-gcn: A temporal graph convolutional network for traffic prediction[J]. IEEE transactions on intelligent transportation systems, 2019, 21(9): 3848-3858.

[17] Li Y, Yu R, Shahabi C, et al. Diffusion convolutional recurrent neural network: Data-driven traffic forecasting. ICRL, 2018, pp. 1-16.

[18] Yu B, Yin H, Zhu Z. Spatio-temporal graph convolutional networks: A deep learning framework for traffic forecasting. IJCAI., 2019, pp. 3634-3640.

[19] Guo S, Lin Y, Feng N, et al. Attention based spatial-temporal graph convolutional networks for traffic flow forecasting. AAAI. 2019, 33(01): 922-929.

[20] Song C, Lin Y, Guo S, et al. Spatial-temporal synchronous graph convolutional networks: A new framework for spatial-temporal network data forecasting. AAAI, 2020, pp.914-921.

[21] Zhang X, Huang C, Xu Y, et al. Traffic flow forecasting with spatial-temporal graph diffusion network. AAAI. 2021, 35(17): 15008-15015.

[22] Li M, Zhu Z. Spatial-temporal fusion graph neural networks for traffic flow forecasting. AAAI. 2021, 35(5): 4189-4196.

[23] Cui Z, Henrickson K, Ke R, et al. Traffic graph convolutional recurrent neural network: A deep learning framework for network-scale traffic learning and forecasting[J]. IEEE Transactions on Intelligent Transportation Systems, 2019, 21(11): 4883-4894.

[24] Fan S, Zhu J, Han X, et al. Metapath-guided heterogeneous graph neural network for intent recommendation. KDD, 2019: 2478-2486.

[25] Liu S, Ounis I, Macdonald C, et al. A heterogeneous graph neural model for cold-start recommendation. SIGIR. 2020: 2029-2032.

[26] Wang D, Liu P, Zheng Y, et al. Heterogeneous graph neural networks for extractive document summarization. ACL, 2020, pp. 6209–6219.

[27] Zhang C, Huang C, Yu L, et al. Camel: Content-aware and meta-path augmented metric learning for author identification. WWW. 2018: 709-718.

[28] Zhang Y, Fan Y, Ye Y, et al. Key player identification in underground forums over attributed heterogeneous information network embedding framework[C]. CIKM. 2019: 549-558.

[29] Fan Y, Zhang Y, Hou S, et al. idev: Enhancing social coding security by cross-platform user identification between github and stack overflow. IJCAI. 2019.

[30] Zhang Y, Wang D, Zhang Y. Neural IR meets graph embedding: A ranking model for product search. WWW, 2019: 2390-2400.

[31] Liu Z, Chen C, Yang X, et al. Heterogeneous graph neural networks for malicious account detection. CIKM. 2018: 2077-2085.

[32] Liu J, Guan W. A summary of traffic flow forecasting methods[J]. Journal of highway and transportation research and development, 2004, 21(3): 82-85.

[33] Wu C H, Ho J M, Lee D T. Travel-time prediction with support vector regression[J]. IEEE transactions on intelligent transportation systems, 2004, 5(4): 276-281.

[34] Cho K, Van Merriënboer B, Bahdanau D, et al. On the properties of neural machine translation: Encoder-decoder approaches. Computer Science, pp.103-111, Oct. 2014.

[35] Wu Z, Pan S, Long G, et al. Graph wavenet for deep spatial-temporal graph modeling. IJCAI. 2019, pp. 1907-1913.

[36] P. Lopez et al. “Microscopic Traffic Simulation using SUMO,” In Proc, 21st IEEE Intell. Transp Syst. Conf. 2018, pp. 2575-2582.

[37] Cantarella, Giulio E., and Ennio Cascetta. "Dynamic processes and equilibrium in transportation networks: towards a unifying theory." Transportation Science 29.4 (1995): 305-329.